\documentclass[conference]{IEEEtran}
\IEEEoverridecommandlockouts
\usepackage{cite}
\usepackage{amsmath,amssymb,amsfonts}
\usepackage{algorithmic}
\usepackage{graphicx}
\usepackage{textcomp}
\usepackage{xcolor}
\def\BibTeX{{\rm B\kern-.05em{\sc i\kern-.025em b}\kern-.08em
    T\kern-.1667em\lower.7ex\hbox{E}\kern-.125emX}}
\begin{document}

\title{Challenges in Designing Teacher Robots with Motivation Based Gestures\\
}
\makeatletter
\newcommand{\linebreakand}{%
  \end{@IEEEauthorhalign}
  \hfill\mbox{}\par
  \mbox{}\hfill\begin{@IEEEauthorhalign}
}
\makeatother

\author{\IEEEauthorblockN{Hae Seon Yun}
\IEEEauthorblockA{\textit{Department of Computer Science} \\
\textit{Humboldt University of Berlin}\\
Berlin, Germany \\
yunhaese@informatik.hu-berlin.de}
\and
\IEEEauthorblockN{Volha Taliaronak}
\IEEEauthorblockA{
\textit{Adaptive Systems Group,} \\ \textit{Department of Computer Science} \\
\textit{Humboldt University of Berlin}\\
Berlin, Germany \\
volha.taliaronak@student.hu-berlin.de}
\and
\IEEEauthorblockN{Murat Kirtay}
\IEEEauthorblockA{
\textit{Adaptive Systems Group,} \\ \textit{Department of Computer Science} \\
\textit{Humboldt University of Berlin}\\
Berlin, Germany \\
murat.kirtay@informatik.hu-berlin.de}
\and
\IEEEauthorblockN{Johann Chevelère}
\IEEEauthorblockA{\textit{Department of Educational Sciences} \\
\textit{University of Potsdam}\\
Potsdam, Germany \\
chevalere@uni-potsdam.de}
\and
\IEEEauthorblockN{Heiko Hübert}
\IEEEauthorblockA{
\textit{School of Engineering}\\
\textit{Energy and Information} \\
\textit{HTW Berlin - University of Applied Sciences}\\
Berlin, Germany \\
Heiko.Huebert@HTW-Berlin.de}
\and
\IEEEauthorblockN{Verena V. Hafner}
\IEEEauthorblockA{
\textit{Adaptive Systems Group,} \\ \textit{Department of Computer Science} \\
\textit{Humboldt University of Berlin}\\
Berlin, Germany \\
hafner@informatik.hu-berlin.de}
\linebreakand
\IEEEauthorblockN{Niels Pinkwart}
\IEEEauthorblockA{\textit{Department of Computer Science} \\
\textit{Humboldt University of Berlin}\\
Berlin, Germany \\
pinkwart@hu-berlin.de}
\and
\IEEEauthorblockN{Rebecca Lazarides}
\IEEEauthorblockA{\textit{Department of Educational Sciences} \\
\textit{University of Potsdam}\\
Potsdam, Germany \\
rebecca.lazarides@uni-potsdam.de}
}
\maketitle

\begin{abstract}
Humanoid robots are increasingly being integrated into learning contexts to assist teaching and learning. However, challenges remain how to design and incorporate such robots in an educational context. As an important part of teaching includes monitoring the motivational and emotional state of the learner and adapting the interaction style and learning content accordingly, in this paper, we discuss the role of gestures displayed by a humanoid robot (i.e., Pepper robot) in a learning and teaching context and present our ongoing research on designing and developing a teacher robot. 
\end{abstract}


\section{Introduction}
Interest in social robots in education is increasing and various research presents positive effects of social robots on learning. For instance, social robots play potent roles in learners’ motivation and engagement that mediates learning gains \cite{1}. This paper aims to understand learners' adaptive motivation and develop artificial learning companions to facilitate learners' emotion and motivation. We base our research and development on an existing intelligent tutoring system named Betty's Brain \cite{betty} which includes two virtual agents (e.g., peer and mentor) to teach a learner a science topic (e.g., climate change) by adopting a ``learning by teaching” paradigm. Our aim is to replace the virtual agents with robots that can exhibit gestures and speech. To design these artificial social learning companions, we focus on the gestures of teachers and on replicating them on the Pepper humanoid robot to assess the motivation of students.
In the following sections, we explain related work that grounds our design decisions and the current state of our work in the context of human-robot interaction.

\section{Background and Related Work}
In \cite{2}, the authors show that people conceive a human-like robot as a teacher and a toy or animal-like robot as a peer. Based on this finding, we replaced the virtual teacher agent in Betty's Brain (Mr. Davis) with a Pepper robot and the peer agent (Betty) with a Cozmo Robot as shown in Figure \ref{fig_betty_cosmo}. Even though the design of a peer agent with a Cozmo Robot is crucial, we focus on presenting our design efforts of teacher robot in this paper.
 
 \begin{figure}[htbp]
\centerline{\includegraphics[width=80mm]{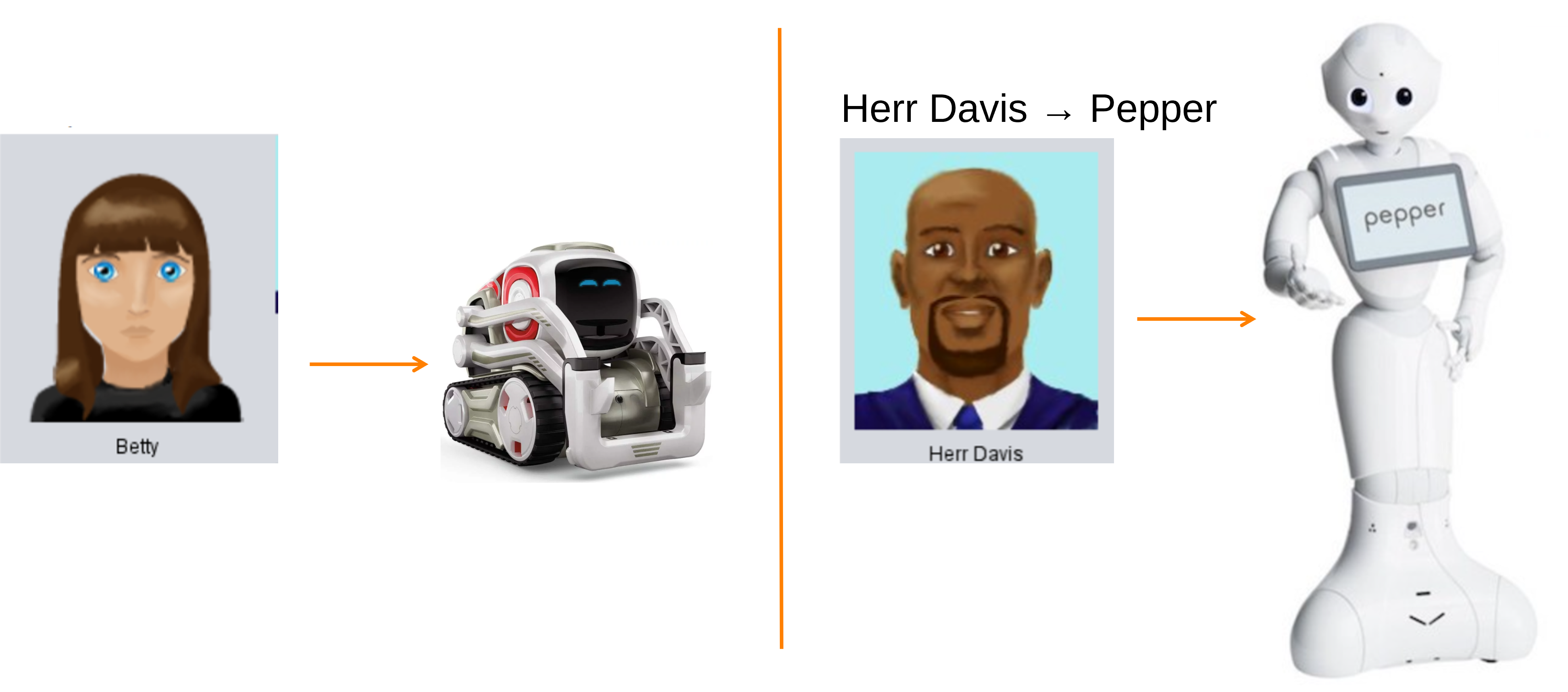}}
\caption{Replacement of virtual agents to robots.}
\label{fig_betty_cosmo}
\end{figure}

Our design decisions of developing the teacher robot are based on the work of \cite{3} which states that teacher’s gestures promote learners’ attention, increase learners’ comprehension, and enhance their learning gains, in addition to promoting learner’s motivation \cite{1}. In the robotic application, the study in \cite{vignolo2019adaptive} advocates that the robotic teacher should incorporate movement to motivate learners to learn effectively with persistence and \cite{5} supports that a robot should incorporate gestures and speech to ensure its livelihood.
To this end, a robotic teacher should be designed as similar as possible to a human  \cite{4},\cite{9}, \cite{14}.

Mirroring a teacher’s gestures to a robotic agent may be a plausible way to assure a human-likeness for a robotic teacher, which in turn promotes an authentic interaction with a learner. The study in \cite{6} shows that human-based contextual gestures were found favorable and considered as human-like. Salem et al. emphasized the importance of acquiring true human interaction and synchronizing between speech and gestures \cite{5}. 
A method to animate a robotic teacher such as Wizard of Oz poses ethical issues of deceiving learners \cite{15} and using an autonomous robot with automatic speech recognition and object detection is not fully reliable \cite{1},\cite{7}. Another approach may be using the pre-recorded human teacher’s gestures and applying them to a robotic teacher. Taking the latter option has the potential to solve the issues of robotic gestures being insensitive to the context and misdirecting learners’ attention discussed in \cite{7}. 
In this study, unlike the above mentioned studies, we thus record the gestures of human teachers using various recording devices and reflect on a robotic teacher that exhibits these gestures.  

\section{Design Decisions}
Design requirements that we elicit in our teacher robot implementation include: 1) the robotic teacher should behave like a human (teacher), 2) the robotic teacher gestures and speech generation should involve open-source offline solutions which do not require data being uploaded and processed online, in addition to being used in various platforms in the future and ensure data privacy, 3) the implementation of a robotic teacher should be sustainable and practical enough so that a teacher without in-depth technical background can utilize it, and 4) human teachers should be involved in a design and development stage to ensure its reliability as a robotic teacher and to increase the ecological validity. 

To design a human-like robotic teacher, we recruited two teachers to record their speech and gestures. To immerse the teachers in a conversational scenario, we engaged them individually in a role-play scenario where they take the Mr. Davis’ role speaking to a student and Betty (e.g., researcher). Both teachers were provided with the scripts of Mr. Davis and they were to read them out loud while moving their arms in front of data recording devices. We have recorded their gestures and speech using a webcam and Kinect V2 camera as shown in
Figure \ref{fig1_cosmo_pepper}. 

\begin{figure}[htbp]
\centerline{\includegraphics[width=80mm]{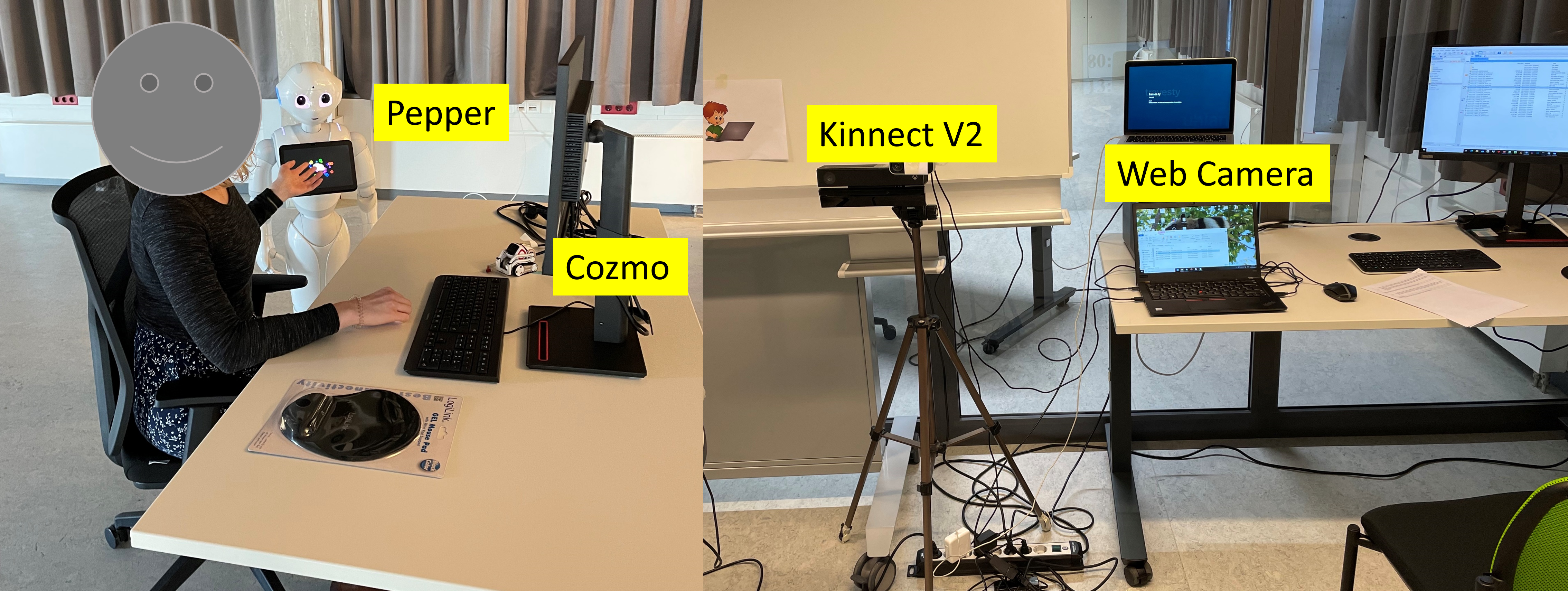}}
\caption{Experimental Setup (left: teacher with Pepper and Cozmo, right: recording devices).}
\label{fig1_cosmo_pepper}
\end{figure}

We employed open-source offline solutions to detect gestures automatically based on the video recordings of two recruited teachers \cite{8}. To compare these gestures with non-recorded alternatives, we enabled the Pepper robot to perform a sequence of movements by using the existing movement library to generate additional versions of a robotic teacher. 
To ensure the reliability and ecological validity of an implemented robotic teacher, we invited six teachers to view five versions of a robotic teacher (three versions using the existing movement library and two versions using the video recordings of teachers) and share their perception on suitability as a teacher and the factors that affect their decision. We note that state-of-the-art surveys, such as  Godspeed, RoSAS \cite{10}, are not succinct and appropriate enough to reflect teachers’ comments on our development of a robotic teacher. Therefore, we have focused on qualitative evaluation similar to a perception study \cite{13} in addition to an interview. 
While the results are currently being analyzed in detail, the preliminary findings support our design decisions on building a robotic teacher that uses gestures based on recordings of human teachers.

\section{Implications and Discussion}\label{AA}
Our research requires reviewing and basing our ideas on theoretical work in addition to considering practical applications and constraints. We established our design of a robotic teacher based on the gestures of a human teacher and utilized an offline open-source software to prospectively empower teachers to design and use teacher robots in their use cases themselves.  

In our setting, we leverage machine learning based solutions to automatically detect the gestures, yet presenting the current state  to a user (in our case, a teacher) still poses a knowledge gap and requires further iterative research. To reduce the technical challenge, we will incorporate speech recognition and further improve the synchronization of gestures and speech. Our future work will continuously involve interdisciplinary researchers ranging from robotics to education, and engage teachers in a co-design process using human-centered design methodology discussed in \cite{bjorling2019participatory} so that our final artifact is not only another technical development to verify research hypothesis and the research findings but also a shared experience increasing the positive effects of research progress and outcomes with the people who are involved. Furthermore, we will conduct a classroom experiment to examine the effect of physical robotic agents on learner's affective and cognitive learning outcomes.   

\section*{Acknowledgment}
This project was funded by the Deutsche Forschungsgemeinschaft (DFG, German Research Foundation) under Germany’s Excellence Strategy – EXC 2002/1 “Science of Intelligence” – project number 390523135.

\bibliographystyle{IEEEtran}
\bibliography{main}
\vspace{12pt}
\end{document}